%% file: main.tex
\documentclass{article} 
\usepackage{iclr2027_conference,times}

\input{math_commands.tex}

\usepackage{graphicx}
\usepackage{hyperref}
\usepackage{url}
\usepackage{float}
\usepackage{rotating}          
\usepackage{graphicx}%
\usepackage{multirow}%
\usepackage{amsmath,amssymb,amsfonts}%
\usepackage{amsthm}%
\usepackage{mathrsfs}%
\usepackage[title]{appendix}%
\usepackage{xcolor}
\usepackage{amssymb} 
\usepackage{textcomp}%
\usepackage{manyfoot}%
\usepackage{booktabs}%
\usepackage{algorithm}%
\usepackage{algorithmicx}%
\usepackage{algpseudocode}%
\usepackage{listings}%
\usepackage{bm}
\usepackage{makecell}
\usepackage{arydshln}

\definecolor{newgreen}{HTML}{08C908}
\newcommand{\yes}{\textcolor{newgreen}{\checkmark}}
\newcommand{\partialyes}{\textcolor{black}{\textbf{$\bm{\circ}$}}}
\newcommand{\no}{\textcolor{red}{$\bm{\times}$}}

\title{Toward a Unified Mathematics of Concepts}

\author{Chen Shani \\
School of Industrial \& Intelligent Systems Engineering\\
Tel Aviv University\\
Israel \\
\texttt{cshani@tauex.tau.ac.il}
}

\begin{document}

\maketitle

\begin{abstract}
Concepts are commonly defined as abstract, compact representations of knowledge and treated as basic units of intelligent behavior. Yet, cognition, psychology, and AI lack a shared mathematical language for them. Modern systems represent concepts as vectors, distributions, symbols, graphs, and other structures, but these formalisms are typically treated as competing rather than as solutions to a common problem. We propose an operation-based view that evaluates mathematical frameworks by the conceptual operations they support, identifying thirteen operations (including similarity, composition, generalization, and grounding) that recur across cognition, psychology, and AI. We show that ten frameworks embody distinct commitments to concepts as self-contained content, relational structure, or evolving process, and that these commitments determine which operations each supports naturally. For example, vector-based models facilitate graded similarity and generalization but struggle with explicit composition, whereas symbolic models support composition but offer but generalize poorly. No single framework we examined naturally supports all operations without extension. 
We test this account empirically using categorization as a case study, operationalizing nine theories on the same items against human judgments. Despite addressing the same conceptual question, the theories produce different procedures and results, demonstrating that mathematical commitment shapes what a theory can explain. We call for hybrid formalisms that treat content, relation, and process as jointly primary.
\end{abstract}

\section{Introduction}\label{sec:intro}

\begin{quotation}
\noindent ``One of the pleasures of looking at the world through mathematical eyes is that you can see certain patterns that would otherwise be hidden.'' \textit{-- Steven Strogatz}
\end{quotation}





\textbf{Concepts are how intelligence compresses the world}. A single concept like \emph{zebra} lets us recognize a novel instance despite changes in pose or lighting, generalize to related categories like \emph{horse}, understand that \emph{horse} + \emph{stripes} = \emph{zebra}, and combine it with others into new thoughts like a \emph{zebra tea party}. Modern LLMs, too, appear to form internal representations that support analogy, generalization, and other documented concept operations. This hints that concepts are as fundamental to artificial systems as to biological ones.

While researchers agree that many cognitive systems converge to similar abstractions (concepts), there is \textbf{no agreed mathematics of concepts} \citep{lawrencec2023natural}. Modern AI represents concepts as vectors, graphs, symbolic predicates, probabilistic variables, sparse features, and many other forms \citep{johnston2023abstract}. These are often viewed as competing representations, each with their own strengths and limitations. We argue that this framing is premature. \textbf{Before asking which mathematics best represents concepts, we must first ask what it is we wish to represent}.

Scientific progress has often been driven by finding the right mathematical language. Differential equations transformed mechanics, probability formalized uncertainty, and graph theory reshaped the study of networks. Mathematics does not merely describe phenomena, but rather determines which structures become natural to represent and which questions become possible to ask. If concepts are the fundamental units of intelligence, then the mathematics used to represent them will shape the kinds of intelligent systems we can imagine and build.

We propose that the fundamental object of study is not a particular representation of concepts, but the capabilities that concepts enable. Intelligence does not merely store concepts; concepts enable intelligent systems to compare and categorize, abstract and generalize, compose and reason, adapt meaning to context, connect knowledge to the world, and acquire and revise conceptual knowledge. At the same time, mathematical frameworks differ in whether they treat concepts as entities with intrinsic structure, as elements defined through their relations, or as processes that emerge and change over time. \textbf{A mathematics of concepts must therefore support the operations concepts support}.

We derive these operations by identifying capabilities that recur independently across comparative cognition, cognitive psychology, and AI, and use them to evaluate existing mathematical approaches. We argue that different mathematical languages capture complementary aspects of conceptual intelligence and together define a new research agenda: the mathematics of concepts.

This paper makes the following contributions: (i) an operation-based framework that defines a mathematics of concepts by the conceptual operations it supports rather than by a preferred representation, identifying thirteen operations that recur across comparative cognition, cognitive psychology, and AI (\S~\ref{sec:def_concepts}); (ii) a systematic comparison of ten mathematical frameworks against these operations, showing that each framework's profile of strengths tracks a prior commitment to content, relation, or process (\S~\ref{sec:content}, \ref{sec:relation}, \ref{sec:dynamic}, Table \ref{tab:framework_comparison}); (iii) an empirical categorization test-case operationalizing nine of these frameworks on the same dataset against human judgments, showing how mathematical commitment shapes both the procedure and the result for a single conceptual question (\S~\ref{sec:categorization}, Table~\ref{tab:empirical_categorization}); and (iv) a diagnosis of why no existing framework spans content, relation, and process jointly, and a concrete agenda for hybrid formalisms that treat all three as jointly primary.

\section{What Should a Mathematics of Concepts Explain?} \label{sec:def_concepts}

\input{gloss}

Every mathematical theory develops around the assumptions and operations it is designed to capture. Thus, we ask: \emph{what do intelligent systems actually do with concepts?} This \emph{operation-based view} characterizes concepts by the cognitive and computational operations they support, rather than by their representations alone. This would define the requirements for any mathematics of concepts.

\subsection{Fundamental Conceptual Operations}

Table~\ref{tab:operations} introduces the 13 concept operations we identified to recur independently across comparative cognition, cognitive psychology, and AI. Although these traditions differ substantially in their methods, assumptions, and explanatory goals, they repeatedly converge on a common set of capabilities that concepts enable across intelligent systems.\footnote{We do not claim that these operations form a complete or final taxonomy.}

Consider the concept \emph{zebra} \includegraphics[height=1.2em]{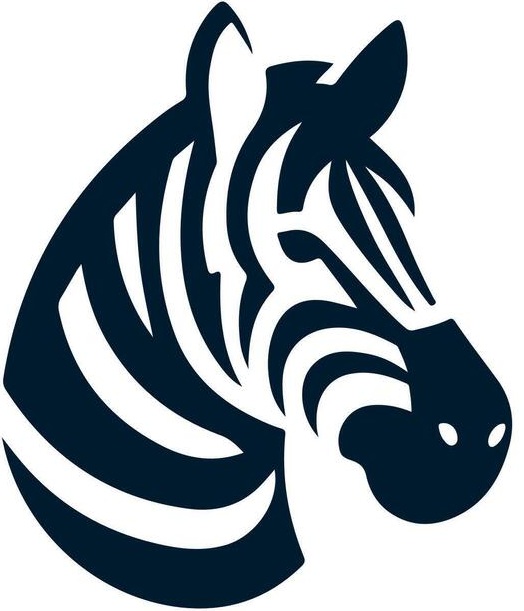}: Abstraction extracts features shared across particular zebras, such as their characteristic stripe pattern, while discarding incidental detail such as which zoo or habitat they are observed in; analogy maps the relation between \emph{horse} and \emph{zebra} onto the relation between \emph{dog} and \emph{wolf}, a domesticated-to-wild correspondence; categorization assigns a new instance to the \emph{zebra} category (e.g., a new zebra at the local zoo); communication allows the word ``zebra'' to reliably evoke the same concept in a listener as in the speaker; composition combines the simpler concepts \emph{horse} and \emph{stripes} into \emph{zebra}, such that the meaning of the whole is predictable from its parts; contextualization changes its interpretation across uses such as an actual \emph{zebra} versus a ``zebra crossing''; discrimination distinguishes \emph{zebra} from the closely confusable concept \emph{horse}; generalization applies \emph{zebra} to a previously unseen subspecies, such as the imperial zebra; grounding connects the concept to sensory experience of actual zebras encountered in the world (the way they feel, smell, etc.); inference derives that a particular zebra is a herbivore from its membership in \emph{zebra}, without observing this directly; learning acquires the concept \emph{zebra} in the first place from repeated exposure to instances, for example through picture books or zoo visits; similarity judges \emph{zebra} as closer to \emph{horse} than to \emph{giraffe}; and revision updates \emph{zebra}'s boundaries on learning that Okapi, despite a very similar stripe pattern, is not a type of \emph{zebra}.

\section{Evaluating Mathematical Frameworks}

The multifaceted nature of concepts suggests that mathematical frameworks will exhibit systematic strengths and weaknesses across conceptual operations. We test this hypothesis by comparing ten existing mathematical languages using the operations framework introduced above. Appendix~\ref{app:frameworks} summarizes each framework's assumptions, mathematical objects, constructions, and philosophy.

But mathematical frameworks do not only differ in \emph{how well} they support a given operation; they differ in \emph{what kind of commitment} they make about a concept in the first place. Some languages treat a concept as a self-contained object, defined by its content independent of other concepts, as in a vector or a probability distribution. Others treat a concept as inherently relational, defined by its connections and compositions with other concepts, as in a graph or a logical predicate. Others define concepts by how they emerge, adapt, or stabilize through interaction, as in a dynamical system or a game-theoretic equilibrium. This reflects a prior choice about the basic unit of analysis, which, in turn, constrains the operations a framework can express support.

This distinction organizes the frameworks into three groups: \emph{content frameworks}, which characterize concepts by their standalone properties (\S~\ref{sec:content}); \emph{relational frameworks}, which characterize concepts by their connections to other concepts (\S~\ref{sec:relation}); and \emph{process frameworks}, which characterize concepts by how they change over time or through interaction (\S~\ref{sec:dynamic}). Comparing frameworks within each group reveals complementary strengths that share a common commitment (e.g., Geometry and Probability both treat concepts as standalone entities, but emphasize different aspects of their content); comparing across groups reveals deeper tradeoffs. Table~\ref{tab:framework_comparison} provides an overview, and the following sections examine each group and framework in turn.

\input{support_tab}

\section{How Are Concepts Represented on Their Own?} \label{sec:content}

Some mathematical languages characterize a concept primarily by what it \emph{is} in isolation. A vector, a probability distribution, or a compressed code all specify a \textbf{concept's content on its own terms}, prior to any explicit connection to other concepts. We group these together as \emph{content frameworks}: geometric, probabilistic, and information-theoretic representations. As we show below, each captures a different notion of conceptual content, be it position, uncertainty, or compression, but all treat the concept as a standalone object rather than a node embedded in a network of relations.

\subsection{Geometric Representations}

Geometric representations, including vector spaces and embedding models, are among the most widely used mathematical languages for representing concepts in modern AI. Their primary strength is expressing concepts through spatial relationships. Similarity, interpolation, clustering, and analogical reasoning emerge naturally from geometric structure, allowing concepts to vary continuously while preserving meaningful neighborhoods. Subspaces further capture shared features and variation, supporting efficient learning and generalization from high-dimensional data.

However, \textbf{geometric structure alone does not specify how conceptual representations are composed, grounded, or manipulated in reasoning}. Vector arithmetic captures simple compositional regularities, such as analogy completion, but more complex composition typically requires learned operations or additional structure. Communication, grounding, and inference likewise are not primitive to geometric representations and must be built into the system. Geometry therefore provides a flexible foundation for conceptual relationships and continuous variation, while leaving the mechanisms of structured reasoning underdetermined.

\subsection{Probabilistic Representations}

Probabilistic frameworks represent concepts through uncertainty. Instead of asking how similar two concepts are, they ask how likely observations, hypotheses, or conceptual assignments are given available evidence. This naturally supports reasoning under uncertainty, evidence accumulation, Bayesian updating, and prediction; operations that are indispensable whenever concepts must be inferred from incomplete or noisy observations, and that probabilistic models handle through principled mechanisms for combining evidence and adapting beliefs as new information arrives.

Their limitations arise elsewhere. \textbf{While probability offers an elegant language for uncertainty, it provides comparatively little structure for expressing compositional relationships or higher-order conceptual organization.} Abstraction and symbolic reasoning must therefore be introduced independently of the probabilistic formalism itself.

\subsection{Information-Theoretic Representations}

Information theory approaches concepts in terms of the information they preserve or discard. Conceptual abstraction becomes a problem of compression: identifying representations that retain behaviorally relevant distinctions while eliminating unnecessary detail.

This perspective naturally supports operations such as abstraction and invariant representation learning, and has become influential in neuroscience and ML, where it provides normative accounts of why intelligent systems develop compressed representations.

Its principal limitation is that \textbf{information theory is largely agnostic about the internal organization of conceptual representations}. It can characterize what information is preserved for reasoning, but does not specify the conceptual relationships or operations needed to compose and manipulate that information. Information theory therefore provides powerful principles for optimizing representations without fully specifying the representations and operations that support reasoning.

\section{How Do Concepts Relate to Other Concepts?} \label{sec:relation}

Other mathematical languages characterize a concept primarily by how it relates to and combines with other concepts. Logical systems, graphs, algebraic structures, and category theory all treat relations, composition, and structure-preserving mappings as the primary explanatory unit, so that a concept matters mainly through its position within a larger relational system. We group these as \emph{relational frameworks}. As we show below, each emphasizes a different kind of relational structure, be it inference, connectivity, closure, or composition, but all share this shift from content to relation.

\subsection{Logical Representations}

Logical systems represent concepts through symbolic predicates, relations, and formal rules. Their strength is explicit inference: deduction, consistency checking, explanation, and rule composition support transparent derivation of complex conclusions.

This explicit structure makes logical representations particularly attractive for domains requiring interpretability, formal guarantees, or precise symbolic reasoning. Unlike geometric approaches, logical derivations can be traced step by step, with each inference interpretable on its own terms.

Their principal weakness is that concepts rarely have purely discrete boundaries. \textbf{Human concepts are graded, context-dependent, and often ambiguous, properties that classical logic does not capture naturally.} Similarity, continuous variation, and statistical learning therefore require substantial extensions beyond the logical formalism itself.

\subsection{Graph-based Representations}

Graphs organize concepts through explicit relationships. Rather than emphasizing individual concepts, they focus on the network of connections among them. Hierarchies, dependencies, taxonomies, and semantic networks therefore emerge naturally within graph structures.

This relational perspective supports operations such as navigation, relational reasoning, and structural organization: traversing paths, computing shortest connections, or aggregating over neighborhoods all fall out naturally from graph structure. Knowledge graphs, ontologies, and semantic networks exemplify this approach across AI and cognitive science.

However, \textbf{graphs alone provide limited mechanisms for representing graded similarity, uncertainty, or abstraction}. While relationships are explicit, the representations themselves generally require complementary mathematical tools to support graded similarity or probabilistic inference.

\subsection{Algebraic Representations}

Algebra characterizes concepts through operations and the rules governing their composition, focusing on the structural properties preserved when concepts are combined or transformed. Mathematical objects such as groups, semigroups, lattices, and Boolean algebras provide formal languages for expressing compositional structure, symmetry, and closure.

This perspective naturally supports operations involving composition, decomposition, symmetry, equivalence, and hierarchical organization. Algebraic methods have long influenced symbolic AI, formal concept analysis, and knowledge representation, offering a principled calculus for building complex conceptual structures from simple ones.

At the same time, \textbf{algebra typically abstracts away continuous variation, uncertainty, and learning dynamics}. While it provides powerful structural constraints, additional mathematical machinery is often required to capture graded similarity, probabilistic reasoning, or adaptation from data.

\subsection{Category-Theoretic Representations}

Category theory treats transformations, rather than objects, as primary, asking how concepts relate through structure-preserving mappings. This makes operations such as composition, abstraction, translation between representations, and structural analogy particularly natural. Universal constructions provide principled notions of abstraction, while functors formalize systematic correspondences between representational systems. In contrast to other frameworks, category theory emphasizes interoperability across mathematical structures.

This generality, however, is also its principal limitation. Category theory provides a language for relating structures rather than prescribing the structures themselves. \textbf{Operations such as similarity, uncertainty, grounding, and learning require additional categorical constructions and domain-specific assumptions to become computationally meaningful.} Category theory is therefore well suited to organizing conceptual operations across otherwise disparate formalisms, but does not by itself provide the content- or process-oriented commitments those formalisms capture.

\section{How Do Concepts Emerge and Evolve?} \label{sec:dynamic}

A third group of mathematical languages characterizes concepts not by their content or relations, but by how they emerge, adapt, and stabilize over time. Dynamical systems, optimization, and game theory all treat concepts as outcomes of an ongoing process, whether driven by temporal dynamics, adaptation to an objective, or strategic interaction among agents, rather than as fixed content or fixed relational structure. We group these as \emph{process frameworks}. Each captures a different driver of conceptual change, but all share this emphasis on concepts as evolving rather than static.

\subsection{Dynamical Systems}

Dynamical systems view concepts not as static representations but as states that evolve over time according to underlying dynamics, asking how conceptual states emerge, stabilize, and transform through interaction with their environment. Such models naturally describe attractors, trajectories, bifurcations, and other temporal phenomena that characterize adaptive behavior.

This perspective makes operations such as concept acquisition, conceptual change, and contextual adaptation particularly natural, and extends naturally to memory (as persistent or recurring states) and sequential reasoning (as trajectories through state space). It has therefore become increasingly influential in neuroscience, cognitive science, and ML, where intelligence is often modeled as a continuously evolving process rather than a sequence of discrete computations.

However, \textbf{dynamical systems provide comparatively little structure for representing explicit symbolic relationships, compositional reasoning, or semantic hierarchies}. They excel at describing \emph{how} conceptual systems evolve, but less naturally express \emph{what} conceptual structures are.

\subsection{Optimization-Based Representations}

Optimization approaches concepts as outcomes of objectives optimized under constraints. Concepts emerge because they support efficient prediction, decision making, communication, compression, or other task goals.

This perspective naturally supports learning, adaptation, generalization, and trade-offs between competing objectives. In modern ML, conceptual representations often emerge implicitly from loss functions, inductive biases, and training dynamics.

\textbf{Its principal limitation is that optimization alone says little about the internal structure of concepts.} Similarity, composition, inference, and abstraction depend on the representational framework being optimized, not on optimization itself. Optimization therefore complements, rather than replaces, frameworks for representing and manipulating concepts.

\subsection{Game-Theoretic Representations}

Game theory studies concepts in the context of strategic interaction among multiple agents. It emphasizes the roles concepts play in collective behavior: how meanings emerge, evolve, and stabilize through communication, cooperation, competition, and negotiation.

This perspective naturally supports operations involving communication, coordination, pragmatic inference, convention formation, and social grounding. It provides a principled framework for understanding how shared conceptual systems arise, adapt, and persist within populations of interacting agents, making it relevant to both human cognition and multi-agent AI.

\textbf{Its principal limitation is its emphasis on interaction rather than internal representation.} Game theory provides limited guidance for representing conceptual structure, similarity, or compositional reasoning within individual agents. It therefore complements representational frameworks by explaining how concepts function and evolve through social interaction.

\section{Categorization Test-Case}
\label{sec:categorization}

To better illustrate how different mathematical frameworks answer questions about concept operations, we ask each framework the same question: \textit{how would this concept be categorized?}. We use a human categorization judgments dataset from \cite{shani2026tokens}. Categorization is a useful test case because almost every framework has something to say about it, yet they view the operation differently. We explore nine frameworks, excluding game theory, which does not naturally support categorization (Table~\ref{tab:framework_comparison}). For each, we implement the corresponding procedure on the same items and compare its output against the same human judgments.

\paragraph{Operationalizations.}
\textit{Geometry} treats a category as a region and clusters by Euclidean distance. \textit{Probability} models each category as a class-conditional density and assigns by posterior. \textit{Optimization} treats categorization as empirical risk minimization and fits a discriminative probe. \textit{Information} uses the Information Bottleneck \citep{tishby2000information}, compressing items while preserving a relevance variable. \textit{Logic} defines a category by features necessary and jointly sufficient for membership. \textit{Algebra} builds the Galois concept lattice over items and their features \citep{ganter1999formal}. \textit{Graph} clusters by shortest-path distance in a knowledge graph; \textit{Structural} clusters by structural equivalence in that graph, discarding node identity. \textit{Dynamical} treats a category as an attractor and measures how stable an item's representation is across training checkpoints. Table~\ref{tab:empirical_categorization} summarizes the operationalizations and results.

\paragraph{Protocol.}
Every method returns a partition of the items, scored against the human partition. When a method admits a choice of cluster number, we set it to the number of human categories, preventing any method from benefiting from its own granularity. Methods that require fitting use nested cross-validation grouped by item, so no item appears in both training and test folds. We report Adjusted Rand Index and Adjusted Mutual Information, both corrected for chance agreement, alongside uncorrected NMI, with confidence intervals obtained by subsampling items, and Accuracy for trained methods with Matched Accuracy for all methods (on the same held-out set). Matched accuracy is cluster accuracy after optimal Hungarian matching to the human categories. Each procedure is tested against a null matched to its confounds: initialization for clustering, checkpoint order for trajectories, and information-budget variation for relevance comparisons. Full methodological details and additional metrics are provided in Appendix~\ref{app:emp}.

\begin{table}[t]
\centering
\small
\setlength{\tabcolsep}{7pt}
\renewcommand{\arraystretch}{1.1}
\begin{tabular}{@{}lcccr@{}}
\toprule
\textbf{Theory} & \textbf{Support} & \textbf{Trained} & \textbf{Accuracy} \\
\midrule
Optimization & \partialyes & Yes & \textbf{0.833} \\
Probability  & \partialyes & Yes & 0.798 \\
\hdashline
Geometry     & \yes        & No  & \textbf{0.607} \\
Graph        & \yes        & No  & 0.552 \\
\hdashline
Logic        & \partialyes & Yes & 0.547 \\
Dynamical    & \partialyes & No  & 0.541 \\
Information  & \partialyes & No  & 0.538 \\
Algebra      & \partialyes & No  & 0.394 \\
Category     & \partialyes & No  & 0.249 \\
\midrule
Game-theory  & \no         & --  & -- \\
\bottomrule
\end{tabular}
\caption{Categorization performance across mathematical frameworks. \textbf{Support} indicates the framework's theoretical support for categorization (Table~\ref{tab:framework_comparison}).}
\end{table}

\subsection{Quantitative interpretation}

The methods differ in ways that cannot be fully equalized. Three of the nine are supervised and use category labels during fitting, so their predictive scores are not directly comparable to the six unsupervised methods. The methods also require different auxiliary resources and therefore cover different subsets of the items: the knowledge graph covers nearly all items, whereas property norms cover 60-76\%. Finally, predictive accuracy is defined only for the supervised methods; for the others, we report cluster accuracy after optimal matching, which measures agreement with the human partition under a different criterion. We therefore do not interpret Table~\ref{tab:empirical_categorization} as a ranking.

These scores indicate how well each operationalization recovers the human category structure under its particular mathematical commitments and input representation. Thus, a low score is evidence that the corresponding operationalization provides a poor account of human categorization in this setting, although it does not by itself identify whether the limitation lies in the mathematical framework, its representation, its auxiliary data, or its implementation. Conversely, higher recovery provides evidence that the framework captures aspects of categorization reflected in the human judgments. 

The two highest-performing methods are supervised, consistent with the advantage of access to category labels during fitting. Among the non-supervised methods, the strongest performers are those classified as \yes\ in Table~\ref{tab:framework_comparison}, suggesting that frameworks that naturally support categorization also tend to provide stronger accounts of human category structure when instantiated on this task.

\subsection{Qualitative interpretation}

We next examine \emph{crowbar}, a Carpenter's Tool in the human data and the item on which the nine mathematical implementations disagree most. It receives seven distinct majority categories, and the average overlap between the sets of items with which it is grouped is only $0.14$ (see Appendix~\ref{app:crowbar}). The four embedding-based accounts place it among birds, the property-norm accounts place it among weapons, and the knowledge-graph account assigns it no coherent neighborhood.

The embedding-based error has a sub-lexical origin. Under mean pooling, $\cos(\emph{crowbar},\emph{crow})=0.735$, and nine of its ten nearest neighbors are birds, while its similarity to other tools does not exceed $0.32$. The compound therefore inherits the neighborhood of its leading subword, an effect also visible in pairs such as \emph{sim}(blackberry, blackjack)=$0.719$. This example also reveals a limitation of apparent convergence across methods. The four embedding-based accounts agree because they share the same input representation and its associated bias, so their agreement is not four independent pieces of evidence. Conversely, disagreement across methods using different representations can reveal which source of evidence drives an error, something an aggregate score obscures.

\section{Conclusions \& Future Work}

We have argued that a mathematics of concepts should be organized around the operations concepts support, such as similarity, composition, generalization, grounding, and revision, rather than around a mathematical framework. Comparing ten mathematical frameworks through this lens shows that each makes different conceptual operations natural, reflecting different commitments to conceptual content, relational structure, and process. No framework in our comparison naturally supports the full set of operations, suggesting that \textbf{a mathematics of concepts will likely require combining complementary mathematical frameworks rather than refining a single one}.

Our categorization experiment provides an initial empirical test of this framework. The results suggest that the theoretical notion of \emph{natural support} is not merely descriptive: among non-supervised methods, the frameworks classified as naturally supporting categorization achieve the strongest recovery of human category structure. At the same time, supervised methods can achieve higher performance by introducing information that is not native to the framework. The results therefore suggest a useful distinction between \emph{native mathematical support} and \emph{performance through added machinery}: a framework's mathematical commitments constrain what it can express naturally, while additional supervision, representations, and auxiliary data can extend what it achieves in practice.

Several directions follow from this framing. First, the empirical analysis should be extended beyond categorization to the broader set of conceptual operations, testing whether the theoretical distinctions in Table~\ref{tab:framework_comparison} predict behavior across tasks and representations. Second, the comparison should be expanded to additional mathematical languages and hybrid formalisms, as well as to a broader and empirically validated taxonomy of conceptual operations. Most importantly, the framework opens a constructive question: \textbf{can we develop mathematical formalisms that support a broad, human-like range of conceptual operations?} We encourage the field to build a mathematical language that makes concepts' full range of operations explicit.





\section*{Reproducibility statement}

We release the code here: \url{https://anonymous.4open.science/r/concept_math-B3BA/README.md}. The data used for the empirical analysis of categorization was published by \citet{shani2026tokens} and can be accessed here: \url{https://huggingface.co/datasets/CShani/human-concepts}.

\bibliography{iclr2027_conference}
\bibliographystyle{iclr2027_conference}

\appendix

\section{Rationale for Table~\ref{tab:framework_comparison}}\label{app:rational}

The assignments in Table~\ref{tab:framework_comparison} are based on whether a mathematical framework provides structures or operations that naturally correspond to each conceptual operation. We use three levels of support: \textit{naturally supported} (\yes) when the operation corresponds to a canonical object or operation of the framework; \textit{supported, but not primary} (\partialyes) when the framework provides relevant mathematical machinery but the operation is not a central object of the framework; and \textit{not naturally supported} (\no) when expressing the operation requires substantial additional assumptions or machinery external to the framework. The rationale for each operation is given below.

\paragraph{Abstraction.}
Abstraction concerns extracting structure that is invariant across particular instances while discarding incidental detail \citep{posner1968genesis}. Information theory (\yes) naturally supports this through representations that retain relevant information while discarding irrelevant variation \citep{cover1991elements}. Algebra (\yes) and category theory (\yes) provide particularly direct structural forms of abstraction: algebra identifies common structure through abstract operations and axioms \citep{repec:spr:sprchp:978-1-4471-0039-3_1, 10.1093/oso/9780190641221.003.0008}, while category theory abstracts away from the internal details of objects in favor of their relationships and mappings \citep{mac1971categories}. Logic (\yes) also supports abstraction through variables, predicates, quantification, and formal structures that separate general properties from particular instances \citep{quine1982methods, 10.1093/oxfordhb/9780199935314.013.004}. Geometry (\partialyes) can support abstraction by identifying invariant geometric structure under transformations \citep{martin2012transformation}. Probability (\partialyes) can likewise support abstraction by identifying probabilistic structure that remains invariant under transformations or symmetries \citep{kallenberg2005probabilistic}. Graphs (\partialyes) provide abstraction by representing entities and their relations independently of their concrete implementation, and by supporting structural representations that preserve relational patterns while discarding incidental detail \citep{battiston2020networks}, while optimization (\partialyes) supports abstraction more indirectly by selecting representations or solutions that preserve task-relevant structure under an explicitly specified objective \citep{tishby2000information}. Dynamical systems (\no) and game-theory (\no) do not naturally support abstraction.

\paragraph{Analogy.}
Analogy concerns mapping structural relationships from one conceptual domain onto another \citep{gentner1983structure}. Geometry (\yes) naturally supports analogy through transformations, isometries, correspondences, and mappings between geometric structures \citep{martin2012transformation}. Category theory (\yes) provides an especially direct formal language for structural correspondence through functors and related structure-preserving mappings \citep{osherson1990category}. Logic (\partialyes), graphs (\partialyes), and algebra (\partialyes) can express structural correspondences between formal systems through interpretations, homomorphisms, and other structure-preserving mappings \citep{enderton2001mathematical, hell2026graphs, repec:spr:sprchp:978-1-4471-0039-3_1}. Probability (\partialyes) and information theory (\partialyes) can support analogy through probabilistic or information-theoretic comparisons between domains \citep{gibbs2002choosing, cover1991elements}, but these are generally secondary uses. Dynamical systems (\no), optimization (\no), and game theory (\no) do not naturally provide cross-domain structural analogy.

\paragraph{Categorization.}
Categorization concerns grouping distinct instances into classes on the basis of shared properties \citep{smith1981categories, murphy1985role, rosch1975family}. Geometry (\yes) naturally supports categorization through clustering, regions, neighborhoods, and geometric structures \citep{mirkin2013mathematical}, and graphs (\yes) support categorization when categories are represented as subgraphs, clusters, or communities \citep{martin2012transformation, fortunato2010community, hell2026graphs} Probability (\partialyes) supports probabilistic class membership and distributions over instances \citep{griffiths2008categorization}; Information theory (\partialyes) can characterize category structure through mutual information shared by instances and category variables \citep{cover1991elements}. Logic (\partialyes) provides predicates that can define classes and formal distinctions, while equivalence relations provide a canonical mathematical mechanism for partitioning objects into classes, although categorization is not itself a central logical operation \citep{enderton2001mathematical}. Algebra (\partialyes) provides equivalence relations, quotient structures, and subalgebras that can partition objects into classes \citep{repec:spr:sprchp:978-1-4471-0039-3_1}; Category theory (\partialyes) organizes objects through morphisms and structural relationships, allowing objects with shared structural properties to be grouped or identified up to isomorphism \citep{mac1971categories}. Dynamical systems (\partialyes) can partition state spaces into regions with qualitatively similar trajectories or asymptotic behavior \citep{strogatz2024nonlinear}. Optimization (\partialyes) can support categorization when category membership is formulated as an optimization or decision problem, but this requires an additional objective or loss function \citep{hart2001pattern}. Game theory (\no) does not naturally provide categorization as a mathematical primitive.

\paragraph{Communication.}
Communication concerns the reliable transmission of a concept between agents \citep{lazaridou2016multi}. Information theory (\yes) provides a way to measure information lose during communication \citep{cover1991elements}. Game theory (\yes) provides canonical models of communication through signaling games, cheap-talk models, and strategic information transmission \citep{crawford1982strategic}. Category theory (\partialyes) can support formal accounts of communication by representing compositional mappings between structured representational systems, although the communicative interpretation requires additional assumptions about encoding and interpretation \citep{coecke2010mathematical}. Dynamical systems (\partialyes) can be adapted such that each system represents an agent's conceptual representation and communication alters the system \citep{strogatz2024nonlinear}. Logic (\partialyes) can represent formal messages, communication protocols, and agents' knowledge states \citep{fagin1995reasoning}, while graphs (\partialyes) can represent communication networks by modeling agents as nodes and communication links as edges, with paths capturing possible routes of information flow \citep{10.1093/acprof:oso/9780199206650.001.0001}. These provide useful machinery, but communication is not itself a primary operation of logic or graph theory. Geometry (\no), probability (\no), and algebra (\no) do not naturally treats inter-agent conceptual transmission as a central operation.

\paragraph{Composition.}
Composition concerns combining simpler concepts into a structured whole according to systematic rules \citep{fodor1988connectionism}. Logic (\yes) provides canonical compositional operations through logical connectives and the recursive construction of well-formed expressions \citep{quine1982methods}. Algebra (\yes) provides composition through binary operations and algebraic structures \citep{repec:spr:sprchp:978-1-4471-0039-3_1}, while category theory (\yes) makes composition of morphisms a foundational operation \citep{mac1971categories}. Graphs (\partialyes) support composition through graph products, substitution, and other graph-combination operations, but composition is not as fundamental as it is in algebra or category theory \citep{hell2026graphs}. Geometry (\partialyes) can represent composite structures through geometric construction and composition of simpler configurations, but composition is secondary to its focus on geometric structure and invariance \citep{borsuk2018foundations}. Probability (\no), information theory (\no), dynamical systems (\no), optimization (\no), and game theory (\no) can be used in compositional models but do not themselves make conceptual composition a primary mathematical operation.

\paragraph{Contextualization.}
Contextualization concerns systematic changes in the meaning or applicability of a concept as surrounding circumstances change \citep{barsalou1982context}. Probability (\yes) naturally supports contextual dependence through conditional distributions and conditional probabilities such as Markov decision processes \citep{jaynes2003probability}. Dynamical systems (\yes) provide a direct representation of state-dependent behavior, in which the current state changes the system's subsequent evolution \citep{strogatz2024nonlinear}. Game theory (\partialyes) represents how an agent's available actions and payoffs depend on the strategic context created by other agents, although contextualized meaning is not its primary object \citep{osborne1994course}. Information theory (\partialyes) can represent contextual dependence directly through conditional entropy, conditional mutual information, and related conditional measures \citep{cover1991elements}. Other frameworks support contextual dependence through additional structural constructions: geometry (\partialyes) through context-dependent metrics or constraints\citep{stillwell2005four}, graphs (\partialyes) through context-dependent relational structure \citep{COURCELLE19923}, category theory (\partialyes) through indexed or fibered constructions \citep{TARLECKI1991239}, and optimization (\partialyes) through context-dependent objectives or constraints \citep{vandenberghe2004convex}. In each case, however, contextual dependence is introduced through an additional construction rather than being a primitive organizing principle of the framework. Logic does not naturally provide context-sensitive conceptual meaning without extending the formalism with an explicit representation of context. Logic (\no) and algebra (\no) do not inherently represent contextual dependence.

\paragraph{Discrimination.}
Discrimination concerns distinguishing between concepts that are close or easily confusable \citep{green1966signal}. Geometry (\yes) naturally supports discrimination through distances, boundaries, and separations in a metric or geometric space \citep{stillwell2005four}. Logic (\yes)provides canonical distinctions through predicates, truth conditions, and entailment \citep{enderton2001mathematical}. Information theory (\yes) provides measures of statistical distinguishability using measures such as Kullback-Leibler and Jensen-Shannon divergence \citep{kullback1951information, lin1991divergence}. Probability (\partialyes) supports discrimination through posterior probabilities and likelihood ratios, which quantify relative support for competing hypotheses \citep{berger1987statistical, casella2024statistical}. Graphs (\partialyes), algebra (\partialyes), category theory (\partialyes), and dynamical systems (\partialyes) can distinguish entities through structural properties, such as graph connectivity, algebraic relations, categorical morphisms and isomorphisms, or differences in dynamical states and trajectories, but they do not intrinsically provide a criterion for deciding which distinctions are relevant for discrimination \citep{west2001introduction, js2022abstract, mac1971categories, strogatz2024nonlinear}. Optimization (\no) and game theory (\no) do not intrinsically provide mechanisms for discrimination; optimization selects solutions according to a specified objective, while game theory characterizes strategic interactions according to specified preferences, actions, and payoffs.

\paragraph{Generalization.}
Generalization concerns extending a concept acquired from limited observations to novel instances \citep{shepard1987toward, mitchell1982generalization}. Probability (\yes) provides canonical machinery for generalization through statistical inference and prediction to unobserved or future cases \citep{casella2024statistical}, while optimization (\yes) provides the objective-based machinery underlying many formal treatments of learning, including the minimization of empirical risk used to study generalization \citep{vapnik2019rethinking}. Geometry (\partialyes) can support generalization through metric neighborhoods, spatial smoothness, interpolation, and geometric decision regions, but these mechanisms do not by themselves specify how experience determines behavior on unseen instances \citep{devroye1996probabilistic}. Information theory (\partialyes) provides tools for characterizing the relationship between learned representations and generalization, but does not itself define a generalization procedure \citep{cover1991elements}. Logic (\partialyes) can support generalization by inducing rules from examples; graphs (\partialyes) by learning reusable relational or local structural patterns; algebra (\partialyes) through reusable algebraic laws and invariants; category theory (\partialyes) through structure-preserving mappings; and dynamical systems (\partialyes) by learning dynamical laws that predict unobserved states or trajectories. In each case, however, an additional learning or inference mechanism is required to determine which structures transfer to unseen instances \citep{hamilton2017inductive, yu2024learning, muggleton1991inductive, repec:spr:sprchp:978-1-4471-0039-3_1, mac1971categories}. Game theory (\no) does not support generalization from one solution to another. 

\paragraph{Grounding.}
Grounding concerns connecting conceptual representations to something external to the conceptual system, such as sensory observations, actions, or the physical environment \citep{harnad1990symbol}. Probability (\yes) naturally supports grounding by linking latent variables or hypotheses to observable evidence through generative and observation models \citep{pearl2014probabilistic}. Graphs (\partialyes) can support grounding by representing relations between abstract concepts and entities or observations in the world, as in knowledge graphs and scene graphs, but grounding requires an additional mapping between graph representations and external or perceptual entities \citep{hofer2024construction}. Information theory (\partialyes) can characterize the statistical relationship between conceptual representations and their external referents through measures such as mutual information, but statistical dependence alone does not establish a grounding or reference relation \citep{usher2001statistical}. Category theory (\partialyes) can represent mappings between conceptual and external domains through morphisms and functors, although the identification and interpretation of those domains must be supplied separately \citep{awodey2006category, mac1971categories}. Dynamical systems (\partialyes) can model coupling between internal states and environmental variables through coupled dynamical equations, although such coupling does not by itself establish a grounding relation \citep{beer2000dynamical}. Optimization (\partialyes) can connect internal representations to actions and environmental outcomes by optimizing task objectives defined over their consequences \citep{sutton1998reinforcement}. Algebra (\no), logic (\no), geometry (\no), and game theory (\no) do not intrinsically provide an external grounding relation. Game theory (\no) can represent interactions with environments or other agents, but interaction alone does not establish semantic grounding.

\paragraph{Inference.}
Inference concerns deriving new beliefs or conclusions from existing conceptual knowledge \citep{osherson1990category}. Logic (\yes) provides canonical machinery for inference through formal deduction and logical entailment \citep{enderton2001mathematical}. Probability (\partialyes) supports probabilistic inference through conditional probability, Bayesian inference, and probabilistic graphical models, but inference depends on a specified probabilistic model and inferential procedure \citep{pearl2014probabilistic}. Graphs (\partialyes) support inference when relational structure encodes dependencies or constraints, allowing information to propagate through paths and connected variables \citep{koller2009probabilistic}. Algebra (\partialyes) supports reasoning through algebraic laws, identities, and structure-preserving transformations, while category theory (\partialyes) supports compositional reasoning through morphism composition, commutative diagrams, and universal properties \citep{js2022abstract, awodey2006category}. Game theory (\partialyes) supports inference about strategic behavior through equilibrium and related solution concepts \citep{fudenberg1991game}. Geometry (\no) and dynamical systems (\no) can support inference about spatial or temporal systems, but it is a different type of inference from the intended one, and optimization (\no) primarily specifies the selection of solutions under an objective and therefore does not, by itself, constitute conceptual inference.

\paragraph{Learning.}
Learning concerns the acquisition of concepts from experience or data \citep{bruner2017study, valiant1984theory}. Probability (\yes) provides a canonical framework for learning through statistical inference and posterior updates \citep{ghahramani2015probabilistic}, and optimization (\yes) provides the central machinery for parameter estimation and empirical risk minimization \citep{vapnik1998statistical}. Information theory (\yes) naturally characterizes learning in terms of information acquisition and better compression \citep{shannon1948mathematical}. Geometry (\yes) can support learning through geometric models of representations and decision boundaries \citep{bronstein2017geometric}. Dynamical systems (\yes) naturally represent learning when it is formulated as a trajectory through parameter or state space \citep{chen2018neural}. Graphs (\partialyes) and category theory (\partialyes) can support learning through graph-structured or relational representations, but learning requires additional algorithms or objectives \citep{battaglia2018relational, coecke2010mathematical}. Logic (\no) and algebra (\no) do not naturally provide mechanisms for acquiring concepts from data.

\paragraph{Similarity.}
Similarity concerns the graded comparison of concepts according to how alike they are \citep{tversky1977features}. Geometry (\yes) provides a canonical account of similarity through metrics, distances, angles, and transformations, and is therefore naturally suited to this operation \citep{do2016differential, stillwell2005four}, while information theory (\partialyes) provides closely related measures of distinguishability such as Kullback-Leibler and Jensen-Shannon divergence \citep{cover1991elements}. Probability (\partialyes) supports graded similarity through likelihoods and distances or divergences between probability distributions \citep{billingsley2012probability}; graphs (\partialyes) can support similarity through graph distances, shared neighborhoods, or structural equivalence \citep{lorrain1971structural}; algebra (\partialyes) through relations or metrics defined on algebraic structures \citep{cohen2016group}; category theory (\partialyes) through structure-preserving mappings \citep{mac1971categories}; and dynamical systems (\partialyes) through distances between states or trajectories \citep{shannon1948mathematical}. In each case, however, similarity is derived from the framework's underlying structure rather than being a primary object of the framework \citep{}. Logic (\no), optimization (\no), and game theory (\no) do not naturally provide graded measures of conceptual similarity.

\paragraph{Revision.}
Revision concerns updating or overturning an existing concept in response to new evidence \citep{alma992590033502466, chi1992conceptual}. Probability (\yes) provides the most direct canonical mechanism through Bayesian updating, in which beliefs are systematically revised in light of observations \citep{jaynes1996probability}. Dynamical systems (\yes) naturally represent revision as state evolution, although the interpretation of the changing state as a revised concept must be supplied \citep{shannon1948mathematical}. Optimization (\partialyes) can support revision through iterative parameter updates, but this depends on an externally specified objective \citep{sutton1998reinforcement}. Graphs (\partialyes), algebra (\partialyes), and category theory (\partialyes) can represent the result of a revision, e.g., an edge or node update in a graph, an algebraic rewriting rule, a new morphism in a category \citep{ehrig2006fundamentals, rozenberg1997handbook, baader1998term, lack2005adhesive}. However, none specifies when incoming evidence should trigger such a change or which of several possible transformations counts as the correct revision, without an externally supplied revision rule. Geometry (\partialyes) can represent movement or deformation of representations, but does not itself provide a mechanism for deciding when an existing concept should be revised \citep{gardenfors2000conceptual}. Logic (\no) can represent nonmonotonic or belief-revision systems, but classical logic alone does not naturally support revision in response to contradictory evidence.

Together, these assessments define the operational comparison in Table~\ref{tab:framework_comparison}. The assignments should not be interpreted as claims about the expressive limits of any framework. Rather, they identify which conceptual operations each mathematical perspective makes most natural. This distinction is important: a framework may be capable of representing an operation without making that operation mathematically primitive. The resulting complementarity suggests that a general mathematics of concepts may be better understood as an integration of mathematical perspectives, each providing different portions of the conceptual interface, than as the search for a single universally sufficient formalism.

\section{Mathematical Frameworks}
\label{app:frameworks}

Table~\ref{tab:frameworks} provides a concise overview of the ten mathematical frameworks considered in our analysis, summarizing their core assumptions, mathematical objects, canonical constructions, and guiding philosophies. We treat these frameworks as distinct mathematical languages for describing conceptual structure: each makes certain properties and operations explicit while leaving others implicit or requiring additional assumptions. The frameworks therefore differ not in whether they can ultimately implement a given conceptual operation, but in how naturally that operation is expressed within their native mathematical structure.

\input{math_frameworks.tex}

\subsection{Empirical Analysis of Categorization}
\label{app:emp}

\input{emp_cat} 

Table~\ref{tab:empirical_categorization} depicts the results of the empirical investigation of the categorization operation of concepts. 

Three of the nine are supervised and use category labels during fitting, so their predictive scores are not directly comparable to the six unsupervised methods. The methods also require different auxiliary resources and therefore cover different subsets of the items: the knowledge graph covers nearly all items, whereas property norms cover 60-76\%. Finally, predictive accuracy is defined only for the supervised methods; for the others, we report cluster accuracy after optimal matching, which measures agreement with the human partition under a different criterion. We therefore do not interpret Table~\ref{tab:empirical_categorization} as a ranking.

These scores indicate how well each operationalization recovers the human category structure under its particular mathematical commitments and input representation. Thus, a low score is evidence that the corresponding operationalization provides a poor account of human categorization in this setting, although it does not by itself identify whether the limitation lies in the mathematical framework, its representation, its auxiliary data, or its implementation. Conversely, higher recovery provides evidence that the framework captures aspects of categorization reflected in the human judgments. 

The two highest-performing methods are supervised, consistent with the advantage of access to category labels during fitting. Among the non-supervised methods, the strongest performers are those classified as \yes\ in Table~\ref{tab:framework_comparison}, suggesting that frameworks that naturally support categorization also tend to provide stronger accounts of human category structure when instantiated on this task.

\subsection{Qualitative Analysis of \emph{crowbar}}
\label{app:crowbar}

Table~\ref{tab:crowbar} provides the full breakdown of the categorizations assigned to \emph{crowbar} by the nine mathematical implementations. In the human data, \emph{crowbar} belongs to the category Carpenter's Tool, but the implementations assign it to seven distinct majority categories. The average overlap between the sets of items with which \emph{crowbar} is grouped is only $0.14$, making it the item with the greatest disagreement across methods.

The disagreement also reveals distinct sources of error. The four embedding-based accounts place \emph{crowbar} among birds. Under mean pooling, $\cos(\emph{crowbar},\emph{crow})=0.735$, and nine of its ten nearest neighbors are birds, while its similarity to other tools does not exceed $0.32$. The compound therefore inherits the neighborhood of its leading subword. A similar effect appears in pairs such as \emph{blackberry}-\emph{blackjack}, which have cosine similarity $0.719$. In contrast, the property-norm accounts place \emph{crowbar} among weapons, while the knowledge-graph account assigns it no coherent neighborhood.

This example illustrates why agreement across implementations should not automatically be interpreted as independent evidence. The four embedding-based accounts share the same input representation and therefore can inherit the same representational bias. Conversely, disagreement between methods based on different representations can help identify which source of information drives an error. Thus, the qualitative analysis complements the aggregate scores by revealing the representational assumptions underlying both convergence and divergence across mathematical accounts.

\input{crowbar} 

\end{document}

%% file: math_commands.tex
\usepackage{amsmath,amsfonts,bm}

\def\eqref#1{equation~\ref{#1}}

\def\1{\bm{1}}

\DeclareMathAlphabet{\mathsfit}{\encodingdefault}{\sfdefault}{m}{sl}
\SetMathAlphabet{\mathsfit}{bold}{\encodingdefault}{\sfdefault}{bx}{n}



%% file: gloss.tex
\begin{table*}[ht!]
\scriptsize
\centering
\begin{tabular}{@{}p{3cm}p{10cm}@{}}
\toprule
\textbf{Operation}  & \textbf{Definition} \\
\midrule

\makecell{\textbf{Abstraction}\\[1.5ex]$A:X^n \rightarrow C$} &
{The capacity to extract structure or regularities that hold across specific instances, discarding incidental detail in favor of what generalizes \cite{posner1968genesis}.}
\\\multicolumn{2}{@{}l}{\textcolor{brown}{{\includegraphics[height=1.2em]{zebra-head.jpg} Extracts shared features, such as zebras’ stripe patterns, while discarding incidental details, such as their habitat.}}}\\
\hline

\makecell{\textbf{Analogy}\\[1.5ex]$A:C_1 \times C_2 \rightarrow R$} &
The capacity to map structural relationships from one conceptual domain onto another, supporting reasoning by structural correspondence rather than surface similarity \cite{gentner1983structure}. 
\\\multicolumn{2}{@{}l}{\textcolor{brown}{{\includegraphics[height=1.2em]{zebra-head.jpg} Maps the relation between HORSE and ZEBRA onto DOG and WOLF, a domesticated-to-wild correspondence.}}}\\
\hline

\makecell{\textbf{Categorization}\\[1.5ex]$K:X \rightarrow C$} &
The capacity to group instances into classes based on shared properties, enabling stable treatment of perceptually or functionally distinct items as members of the same kind \cite{smith1981categories, murphy1985role, rosch1975family}. 
\\\multicolumn{2}{@{}l}{\textcolor{brown}{{\includegraphics[height=1.2em]{zebra-head.jpg} Assigns a new instance to the ZEBRA category (e.g., a new zebra at the local zoo).}}}\\
\hline

\makecell{\textbf{Communication}\\[1.5ex]$M:C \times \mathcal{A} \rightarrow C$} &
The capacity for a concept to be reliably transmitted between agents, supporting shared reference and coordinated use \cite{lazaridou2016multi}. \\
\multicolumn{2}{@{}l}{\textcolor{brown}{{\includegraphics[height=1.2em]{zebra-head.jpg} Allows the word ``zebra'' to reliably evoke the same concept in a listener as in the speaker.}}}\\
\hline

\makecell{\textbf{Composition}\\[1.5ex]$\operatorname{Comp}:C^n \times R \rightarrow C$} &
The capacity to combine simpler concepts into more complex ones according to systematic rules, such that the meaning of the whole depends predictably on its parts \cite{fodor1988connectionism}. Note that unlike Analogy, which is about mapping structure \textit{across} domains, Composition is \textit{within} a domain.\\
\multicolumn{2}{@{}l}{\textcolor{brown}{{\includegraphics[height=1.2em]{zebra-head.jpg} Combines the simpler concepts HORSE and STRIPES into ZEBRA, such that the meaning of the whole is predictable from its parts.}}}\\
\hline

\makecell{\textbf{Contextualization}\\[1.5ex]$Q:C \times X \rightarrow C$} &
The capacity for a concept's meaning or applicability to shift systematically depending on the surrounding context \cite{barsalou1982context}.  \\
\multicolumn{2}{@{}l}{\textcolor{brown}{{\includegraphics[height=1.2em]{zebra-head.jpg} Changes its interpretation across uses such as an actual ZEBRA versus a ``zebra crossing''.}}}\\
\hline

\makecell{\textbf{Discrimination}\\[1.5ex]$D:X \times C \times C \rightarrow \{0,1\}$} &
The capacity to distinguish between concepts that are close or easily confusable, complementary to categorization in that it draws boundaries rather than groupings \cite{green1966signal}. Note that while Similarity, Categorization, and Discrimination all concern how a concept relates to nearby concepts, they differ in whether the operation groups instances together, draws a boundary between them, or yields a graded judgment of closeness. Note that a system can \textit{categorize} broadly without discriminating finely, or discriminate without producing a full partition. \\
\multicolumn{2}{@{}l}{\textcolor{brown}{{\includegraphics[height=1.2em]{zebra-head.jpg} Distinguishes ZEBRA from the closely confusable concept HORSE.}}}\\
\hline

\makecell{\textbf{Generalization}\\[1.5ex]$G:C \times X_{\mathrm{new}} \rightarrow \{0,1\}$} &
The capacity to extend a concept learned from limited instances to novel instances not previously observed \cite{shepard1987toward, mitchell1982generalization}. Unlike Abstraction, which transforms a set of particular instances into a representation that preserves selected invariant structure across them, Generalization extends an already acquired concept to new instances. \\
\multicolumn{2}{@{}l}{\textcolor{brown}{{\includegraphics[height=1.2em]{zebra-head.jpg} Applies ZEBRA to a previously unseen subspecies, such as the imperial zebra.}}}\\
\hline

\makecell{\textbf{Grounding}\\[1.5ex]$R:C \times E \rightarrow [0,1]$} &
The capacity to connect a concept to something outside the conceptual system itself, such as sensory input, embodied action, or the external world \cite{harnad1990symbol}. Note that unlike Contextualization, which concerns how a concept's meaning shifts \textit{between uses}, Grounding concerns whether a concept connects \textit{at all} to something outside the conceptual system, such as perception or embodied action. \\
\multicolumn{2}{@{}l}{\textcolor{brown}{{\includegraphics[height=1.2em]{zebra-head.jpg} Connects the concept to sensory experience of actual zebras encountered in the world (the way they feel, smell, etc.).}}}\\
\hline

\makecell{\textbf{Inference}\\[1.5ex]$I:K \rightarrow K$} &
The capacity to derive new conceptual knowledge from existing knowledge, whether through logical entailment, probabilistic reasoning, or other systematic derivation \cite{osherson1990category}. Note that unlike Composition, which builds a new \textit{concept} out of parts, Inference derives a new \textit{belief or conclusion} from existing conceptual knowledge, without necessarily constructing any new composite concept. \\
\multicolumn{2}{@{}l}{\textcolor{brown}{{\includegraphics[height=1.2em]{zebra-head.jpg} Derives that a particular zebra is a herbivore from its membership in ZEBRA, without observing this directly.}}}\\
\hline

\makecell{\textbf{Learning}\\[1.5ex]$L:\mathcal{D} \rightarrow C$} &
The capacity to acquire new concepts from experience or data, rather than having them specified in advance \cite{bruner2017study, valiant1984theory}. Note that unlike Generalization, which concerns extending an \textit{already-acquired} concept to novel instances, Learning concerns the acquisition of the concept itself. \\
\multicolumn{2}{@{}l}{\textcolor{brown}{{\includegraphics[height=1.2em]{zebra-head.jpg} Acquires the concept ZEBRA in the first place from repeated exposure to instances, for example through picture books or zoo visits.}}}\\
\hline

\makecell{\textbf{Similarity}\\[1.5ex]$S:C \times C \rightarrow [0,1]$} &
The capacity to judge how alike two concepts are, typically supporting graded rather than binary comparison \cite{tversky1977features}.  \\
\multicolumn{2}{@{}l}{\textcolor{brown}{{\includegraphics[height=1.2em]{zebra-head.jpg} Judging ZEBRA as closer to HORSE than to GIRAFFE.}}}\\
\hline

\makecell{\textbf{Revision}\\[1.5ex]$V:C \times E \rightarrow C$} &
The capacity to update or overturn an existing concept when it proves inconsistent with new evidence, as distinct from acquiring an entirely new concept \cite{alma992590033502466, chi1992conceptual}. \\
\multicolumn{2}{@{}l}{\textcolor{brown}{{\includegraphics[height=1.2em]{zebra-head.jpg} Updates the boundaries of ZEBRA on learning that Okapi, despite a very similar stripe pattern, is not a type of ZEBRA.}}}\\

\bottomrule
\end{tabular}
\caption{The thirteen conceptual operations identified across comparative cognition, cognitive psychology, and AI, used as a representation-independent interface for evaluating mathematical frameworks in Table~\ref{tab:framework_comparison}. The order is alphabetical to avoid any claims about importance. {Brown text explains what the operation means using the running ZEBRA concept example.}}
\label{tab:operations}
\end{table*}

%% file: support_tab.tex
\begin{table*}[h!]
\centering
\begin{tabular}{l ccc ccccc cc}
\toprule
\textbf{Operation} &
\rotatebox{90}{Geometry} &
\rotatebox{90}{Probability} &
\rotatebox{90}{Information} &
\rotatebox{90}{Logic} &
\rotatebox{90}{Graphs} &
\rotatebox{90}{Algebra} &
\rotatebox{90}{Category} &
\rotatebox{90}{Dynamical} &
\rotatebox{90}{Optimization} &
\rotatebox{90}{Game-Theory} \\
\cmidrule(lr){2-4} \cmidrule(lr){5-8} \cmidrule(lr){9-11}
& \multicolumn{3}{c}{\textit{Content (\S~\ref{sec:content})}} & \multicolumn{4}{c}{\textit{Relational (\S~\ref{sec:relation})}} & \multicolumn{3}{c}{\textit{Process (\S~\ref{sec:dynamic})}} \\
\midrule

Abstraction       & \partialyes & \partialyes & \yes & \yes & \partialyes & \yes & \yes & \no & \partialyes & \no \\
Analogy           & \yes & \partialyes & \partialyes & \partialyes & \partialyes & \partialyes & \yes & \no & \no & \no \\
Categorization    & \yes & \partialyes & \partialyes & \partialyes & \yes & \partialyes & \partialyes & \partialyes & \partialyes & \no \\
Communication     & \no & \no & \yes & \partialyes & \partialyes & \no & \partialyes & \partialyes & \no & \yes \\
Composition       & \partialyes & \no & \no & \yes & \partialyes & \yes & \yes & \no & \no & \no \\
Contextualization & \partialyes & \yes & \partialyes & \no & \partialyes & \no & \partialyes & \yes & \partialyes & \partialyes  \\
Discrimination    & \yes & \partialyes & \yes & \yes & \partialyes & \partialyes & \partialyes & \partialyes & \no & \no \\
Generalization    & \partialyes & \yes & \partialyes & \partialyes & \partialyes & \partialyes & \partialyes & \partialyes & \yes & \no \\
Grounding         & \no & \yes & \partialyes & \no & \partialyes & \no & \partialyes & \partialyes & \partialyes & \no \\
Inference         & \no & \partialyes & \no & \yes & \partialyes & \partialyes & \partialyes & \no & \no & \partialyes \\
Learning          & \yes & \yes & \yes & \no & \partialyes & \no & \partialyes & \yes & \yes & \partialyes \\
Similarity        & \yes & \partialyes & \partialyes & \no & \partialyes & \partialyes & \partialyes & \partialyes & \no & \no \\
Revision & \partialyes & \yes & \partialyes & \no & \partialyes & \partialyes & \partialyes & \yes & \partialyes & \partialyes \\
\bottomrule
\end{tabular}

\vspace{0.5em}

\yes\   Naturally supported \ \ \ \ 
\partialyes \  Supported, but not primary \ \ \ \ 
\no \ Not naturally supported 

\vspace{0.5em}

\caption{Comparison of mathematical frameworks according to the conceptual operations they naturally support, grouped by content, relational, and process frameworks. The operational perspective reveals complementary strengths rather than a single universally superior mathematical language. A framework naturally supports an operation when the operation is represented by a canonical construction with minimal additional representational assumptions. This does not imply that other frameworks cannot implement the operation. See explanation of the rankings in Appendix~\ref{app:rational}. These symbols can be interpreted as: \yes = Primitive; the operation is a canonical object or construction, \partialyes = Induced; the operation can be derived from standard structures, and \no = Added; the operation requires an external representation, objective, or semantics. The classifications indicate how directly each framework supports an operation, not whether the operation is possible within that framework. No operation is exclusive to a single mathematical language; even when a framework does not support an operation natively, it may implement it through additional structures or assumptions.}
\label{tab:framework_comparison}
\end{table*}

%% file: math_frameworks.tex
\begin{sidewaystable*}[ht!]
\scriptsize
\centering
\begin{tabular}{@{}p{1.5cm}p{5.0cm}p{4.2cm}p{4.2cm}p{6cm}@{}}
\toprule
\textbf{Framework} &
\textbf{Core assumption} &
\textbf{Mathematical objects} &
\textbf{Canonical constructions} &
\textbf{Guiding philosophy} \\
\midrule

\textbf{Geometry} &
Concepts occupy structured spaces in which spatial relationships encode meaningful relationships. &
Points, vectors, spaces, distances, subspaces &
Distance, similarity, projection, interpolation, vector arithmetic, transformations &
Understand concepts through their position and structure in a space: what matters is where concepts lie, how far apart they are, and how they can be transformed or combined geometrically. \\

\hline

\textbf{Probability} &
Conceptual knowledge is inherently uncertain, and representations can be understood as distributions over possible states or interpretations. &
Random variables, probability distributions, conditional distributions &
Conditioning, marginalization, Bayesian updating, expectation, sampling &
Represent what is known not as a fixed value but as a distribution over possibilities, and update that distribution as evidence arrives. \\

\hline

\textbf{Information} &
Representations differ in how much relevant information they preserve, discard, or share. &
Entropy, mutual information, information measures &
Compression, information bottlenecks, information maximization, rate--distortion &
Characterize a good representation by the information it retains about what matters while discarding irrelevant detail. \\

\hline

\textbf{Logic} &
Conceptual knowledge can be expressed through explicit propositions and rules whose validity is determined by formal structure. &
Symbols, predicates, propositions, rules, logical structures &
Deduction, implication, substitution, unification, rule application, logical composition &
Make conceptual structure explicit and derive consequences through formally specified rules, rather than relying on graded similarity alone. \\

\hline

\textbf{Graphs} &
Concepts are entities whose meaning and behavior depend on explicit relationships to other entities. &
Nodes, edges, paths, graphs, networks &
Traversal, connectivity, path finding, neighborhood operations, graph composition &
Understand a conceptual system through its network of entities and relations: what matters is which concepts are connected, how, and through what paths. \\

\hline

\textbf{Algebra} &
Conceptual systems can be characterized by the operations and transformations that act on their elements. &
Elements, operators, groups, algebras, transformations &
Composition, inversion, transformation, equivalence, homomorphism &
Understand concepts through what can be done with them and how operations compose, rather than primarily through their absolute representation. \\

\hline

\textbf{Category} &
The structure of a system is determined by objects and the mappings between them, with emphasis on relationships and compositionality. &
Objects, morphisms, categories, functors &
Composition of morphisms, identity, functors, natural transformations &
Focus on structure-preserving mappings and how transformations compose, treating relationships between representations as fundamental. \\

\hline

\textbf{Dynamical} &
Conceptual states change over time according to systematic transition rules or underlying dynamics. &
States, trajectories, transition functions, dynamical systems &
State transitions, trajectories, attractors, stability, temporal composition &
Understand concepts and representations as processes rather than static objects, emphasizing trajectories, transitions, stability, and change. \\

\hline

\textbf{Optimization} &
Behavior or representations can be characterized as solutions to objectives subject to constraints and trade-offs. &
Objectives, constraints, feasible sets, decision variables &
Minimization/maximization, constraint satisfaction, regularization, gradient-based optimization &
Explain conceptual structure in terms of what a system is optimizing and the constraints under which it operates. \\

\hline

\textbf{Game Theory} &
Conceptual behavior emerges through strategic interaction among agents with potentially different goals and information. &
Agents, strategies, utilities, payoffs, equilibria &
Best response, strategic interaction, equilibrium, coordination, incentive optimization &
Understand conceptual behavior through incentives and strategic dependencies: what an agent does depends on what other agents do and what outcomes they seek. \\

\bottomrule
\end{tabular}
\caption{Conceptual overview of the ten mathematical frameworks considered in our comparison. For each framework, we summarize its core assumption, basic mathematical objects, canonical constructions, and guiding philosophy. These descriptions provide the interpretive assumptions underlying the operational comparison in Table~\ref{tab:framework_comparison}; they are intentionally high-level and do not exhaust the mathematical variants or applications of any framework.}
\label{tab:frameworks}
\end{sidewaystable*}

%% file: emp_cat.tex
\begin{sidewaystable*}[p]
\centering
\footnotesize
\setlength{\tabcolsep}{4.5pt}
\renewcommand{\arraystretch}{1.2}
\begin{tabular}{@{}lccl p{5.0cm} rrrrr@{}}
\toprule
\textbf{Theory} & \textbf{Sup.} & \textbf{Trained} & \textbf{Additional Data} & \textbf{Methodology}  & \textbf{ARI} & \textbf{AMI} & \textbf{NMI} & \shortstack{\textbf{Matched}\\\textbf{Acc.}} & \textbf{Acc.} \\
\midrule
Optimization & \partialyes & Yes & Embeddings
  & Logistic-regression probe (multinomial cross-entropy + L2) by ERM; nested grouped CV
  & {0.717} & {0.767} & {0.793} & {0.833} & {0.833} \\
Probability & \partialyes & Yes & Embeddings
  & Gaussian class-conditionals with shared covariance (LDA, Ledoit--Wolf shrinkage) on PCA; posterior $P(\mathrm{cat}\mid\mathrm{item})$
  & 0.672 & 0.726 & 0.759 & 0.801 & 0.798 \\
Geometry & \yes & No & Embeddings
  & $k$-means (Euclidean, \texttt{n\_init}${=}10$) at $k$~=~number of human categories
  & 0.400 & 0.591 & 0.629 & 0.607 & -- \\
Graph & \yes & No & KG (WordNet)
  & $k$-medoids on WordNet shortest-path distance (max over senses)
  & 0.369 & 0.528 & 0.569 & 0.552 & -- \\
Information & \partialyes & No & Embeddings and KG (WordNet)
  & Agglomerative Information Bottleneck: greedy merges minimising loss of $I(T;Y)$
  & 0.363 & 0.483 & 0.529 & 0.538 & -- \\
Dynamical & \partialyes & No & Embeddings over all checkpoints
  & $k$-medoids on rank-normalised cosine distance averaged across 56 checkpoints; settling from nearest-neighbour sets
  & 0.355 & 0.466 & 0.514 & 0.541 & -- \\
Logic & \partialyes & Yes & Feature norms
  & Per-category majority-vote necessary/sufficient feature rules; nested grouped CV
  & 0.335 & 0.460 & 0.546 & 0.560 & 0.547 \\
Algebra & \partialyes & No & Feature norms
  & Formal Concept Analysis; $k$-medoids on Galois-closure distance
  & 0.193 & 0.297 & 0.419 & 0.394 & -- \\
Category & \partialyes & No & KG (WordNet)
  & Weisfeiler--Lehman refinement (graded bisimulation) on typed WordNet graph; $k$-medoids on WL-kernel distance
  & 0.062 & 0.149 & 0.220 & 0.249 & -- \\
\midrule
Game-theory & \no & -- & -- & -- & -- & -- & -- & -- & -- \\
\bottomrule
\end{tabular}
\caption{Mathematical theories of categorization evaluated against human category judgements. \textbf{Sup.}: \yes=supported by the plan, \partialyes=requires some additional data or training, \no=not testable. Scores are means across the three human datasets weighted by item count; item counts differ by theory (952 for embedding-based theories, 938 for WordNet-based, 722 for Logic and 569 for Algebra, reflecting coverage of the required resource). \textbf{Matched Acc.} is Hungarian-matched cluster accuracy and is available for all theories; \textbf{Acc.} is held-out predictive accuracy and is defined only for the trained theories. The two are not comparable. ARI and AMI are chance-corrected; NMI is not and rises with cluster count. Embeddings were mean pooled extrected from the 24th layer of the final checkpoints using OLMo-7B.}
\label{tab:empirical_categorization}
\end{sidewaystable*}

%% file: crowbar.tex
\begin{table}[t]
\centering
\footnotesize
\begin{tabular}{@{}ll p{4.3cm}@{}}
\toprule
Method & Places \emph{crowbar} with & Representative co-members \\
\midrule
\multicolumn{3}{@{}l}{\emph{Human ground truth: Carpenter's Tool}} \\
\midrule
Optimization & Bird & crane, robin, sparrow, bluejay \\
Probability & Bird & robin, sparrow, bluejay, dove \\
Geometry & Birds & crane, robin, sparrow, canary \\
Dynamical & Birds & (bird cluster, 48/65) \\
Logic & Weapon & gun, pistol, revolver, knife, dagger \\
Graph & Clothing & piano, cushion, cupboard, mirror \\
Algebra & Toy & table, bench, cushion, fan \\
Category & --- & singleton; no co-members \\
Information & Carpenter's tool & stereo, television, radio, telephone \\
\bottomrule
\end{tabular}
\caption{Where each account places \emph{crowbar}.}
\label{tab:crowbar}
\end{table}